\documentclass{article}

\usepackage[accepted]{icml2014}

\usepackage{graphicx}
\usepackage{amsmath,amssymb} 
\usepackage{color}
\usepackage{lscape}
\usepackage{multirow}
\usepackage{url}
\usepackage{hyperref}

\usepackage{times}
\usepackage{epsfig}
\usepackage{gensymb}

\usepackage{xspace}

\usepackage{subfigure}
\usepackage{booktabs}
\usepackage{array}
\makeatletter

\newcommand{\thickhline}{%
    \noalign {\ifnum 0=`}\fi \hrule height 1pt
    \futurelet \reserved@a \@xhline
}

\usepackage{wrapfig}

\definecolor{lightgray}{gray}{.80}

\icmltitlerunning{Can We Teach Computers to Understand Art? }

\begin{document}

\twocolumn[

\icmltitle{Can We Teach Computers to Understand Art? Domain Adaptation for Enhancing
    Deep Networks Capacity to De-Abstract Art    }

\icmlauthor{Mihai Badea}{mbadea@imag.pub.ro} \icmladdress{Image Processing and Analysis Laboratory\\
                University "Politehnica" of Bucharest, Romania,
                Address Splaiul Independen\c{t}ei 313}
\icmlauthor{Corneliu Florea}{corneliu.florea@upb.ro} \icmladdress{Image Processing and Analysis Laboratory\\
                University "Politehnica" of Bucharest, Romania,
                Address Splaiul Independen\c{t}ei 313}

\icmlauthor{Laura Florea}{laura.florea@upb.ro} \icmladdress{Image Processing and Analysis Laboratory\\
                University "Politehnica" of Bucharest, Romania,
                Address Splaiul Independen\c{t}ei 313}
\icmlauthor{Constantin Vertan}{constantin.vertan@upb.ro} \icmladdress{Image Processing and Analysis Laboratory\\
                University "Politehnica" of Bucharest, Romania,
                Address Splaiul Independen\c{t}ei 313}

\icmlkeywords{Convolutional Neural Networks, Domain adaptation, Genre recognition, Painting
analysis, Style transfer.
}

\vskip 0.3in

]

\begin{abstract}
Humans comprehend a natural scene at a single glance; painters and other visual artists, through
their abstract representations, stressed this capacity to the limit. The performance of computer
vision solutions matched that of humans in many problems of visual recognition. In this paper we
address the problem of recognizing the genre (subject) in digitized paintings using Convolutional
Neural Networks (CNN) as part of the more general dealing with abstract and/or artistic
representation of scenes. Initially we establish the state of the art performance by training a CNN
from scratch. In the next level of evaluation, we identify aspects that hinder the CNNs'
recognition, such as artistic abstraction. Further, we test various domain adaptation methods that
could enhance the subject recognition capabilities of the CNNs. The evaluation is performed on a
database of 80,000 annotated digitized paintings, which is tentatively extended with artistic
photographs, either original or stylized, in order to emulate artistic representations.
Surprisingly, the most efficient domain adaptation is not the neural style transfer. Finally, the
paper provides an experiment-based assessment of the abstraction level that CNNs are able to
achieve.
\end{abstract}

\section{Introduction}
\label{sec:introduction}

This paper aims to investigate the differences between the level of abstraction achieved by deep
convolutional neural networks as compared to the human performance in the context of paining
analysis. To synthesize the motivation, let us recall Pablo Picasso words: ''There is no abstract
art. You must always start with something. Afterward you can remove all traces of reality''. Art
historians and enthusiasts are able to note, while recalling major artistic works through the
history, that the level of abstraction steadily increased.

In parallel, in the last period, works that use computer vision techniques to analyze visual art
increased with respect to both the quantity and the quality of reported results. Two trends favored
these developments. First, there were consistent efforts to digitize more and more paintings, such
that modern systems may learn from large databases. Two of such popular efforts are Your Paintings
(now Art UK\footnote{\url{http://artuk.org/}}) which contains more than 200,000 paintings tightly
connected with historical British culture and WikiArt\footnote{\url{http://www.wikiart.org/}} which
contains around 100,000 paintings gathered from multiple national cultures. The databases come with
multiple annotations. For this work we are particulary interested in annotations dealing with the
painting's subject or scene type. From this point of view, a more complete database is the WikiArt
collection, where the labelling category is named \emph{genre}. The second trend is purely
technical and it deals with the development of the Deep Neural Networks, that allowed
classification performances that were not imagined before. In this work, we will use the more
popular Convolutional Neural Networks (CNN) to recognize the painting genre.

Let us, now, to establish the meaning of genre the relation with scene and with the image subject.
A list of definitions for various paintings genres is presented in Table \ref{Tab:Genre_Explain}.
To label a painting into a specific genre, in most of the cases, a user has to identify the subject
of that painting. The exceptions are ``Abstract Art'', ``Design'', ``Illustration'' and ``Sketch
and Study'', where the main characteristic is related to the depiction mode. In this majority of
cases, subject is related to the scene represented in the work of art. The term ``genre'' is
typical for art domain, and is a more general, including, concept than mere ``subject'' or ''scene
type''. In this work, while referring to paintings, we will use all three with the same meaning of
''genre''. In comparison, for a non-artistic photograph, as there is no artistic intervention in
the depiction mode, the subject is more related to the scene, while the genre is hard to be
defined. For artistic photos, the ``genre'' gets meaning again.

Starting from the idea that Deep Neural Networks share similarities with the human vision
\cite{Cichy16} and the fact that such networks are already proven to be efficient in other
perception-inspired areas, like object recognition or even in creating artistic images, we ask
ourselves if they can pass the abstraction limit of artistic paintings and correctly recognize the
scene type of such a work.

In this paper we will first work with the Residual Network (ResNet) on the standard WikiArt
database so to obtain state of the art results. Afterwards, we will test different domain transfer
augmentations to see if they can increase the recognition rate;  also we will study if the network
is capable to pass the abstraction limit and learn from different types of images that contain the
same type of scenes. Furthermore, we introduce several alternatives for domain transfer to achieve
a dual-task: improve the scene recognition performance and understand the abstraction capabilities
of machine learning systems.

Regarding deep networks, multiple improvements have been proposed. In many situations, if the given
task database is small, better performance is reachable if the network parameters are previously
trained for a different task on a large database, such as ImageNet. Next, these values are updated
to the given task. This is called fine--tuning and it is a case of transfer learning. As our
investigation is related to a different domain transfer, we will avoid to use both of them
simultaneuosly, in order to establish clearer conclusions. To compensate, we are relying on the
recent architecture of the Residual Networks (Resnet \cite{He2016}) that was shown to be able to
overcome the problem of vanishing gradients, reaching better accuracy for the same number of
parameters, when compared to previous architectures.


\subsection{Contribution and paper organization}
This paper extends our previous works \cite{florea2017:Scia,badea2017}, being mostly developed from
\cite{florea2017:Scia}, where we had initiated the discussion about the efficiency of various
methods to transfer information from the photographic domain to the paintings domain, such that the
recognition by CNNs' of paintings genre is improved. In this paper we significantly extend the
discussion, by including other transfer methods and by adding more significant results that allow
crisper conclusions. In the second work (\cite{badea2017}), we showed that the artistic style
transfer remains as efficient even if a reduced number of iterations are performed while
over--imposing the style of an artistic painting and the content from a photograph onto a new
image, according to the neural style transfer introduced by Gatys et al. \citeyear{Gatys:2015}.

Overall, this paper claims several contributions along a number of directions. On one direction, we
investigate which aspects, comprehensible by humans, hinder the CNNs while understanding a painting
genre; subsequently by means of domain transfer, we retrieve information about the internal
description and the organization of the painting clusters. In order to accomplish such a task, we
annotate artistic photographic images with respect to the scene type related to genres and we
stylize a large corpus of photographs using different style transfer methods. All this data will be
made publicly available to be used in other research works.

On a second direction, this paper is the first to objectively evaluate the efficiency of the
currently popular neural style transfer  methods. Currently existing solutions
\cite{Gatys:2015,johnson2016,ulyanov2016,huang2017} compare themselves by speed, stability within
video sequences or number of transferable styles. By quantifying the improvement while adapting
photographs to the painting domain, we reach a surprising conclusion, namely that they are less or
at most as efficient as non-neural style transfers  solutions. Evermore, a CNN finds as informative
the original photographs without any style transfer applied.

The remainder of the paper is organized as follows: section \ref{Sect:Related} presents previous
relevant works, section \ref{Sect:CNN} summarizes the CNN choices made and section
\ref{Sect:PaintingUnderstanding} will discuss different aspects of painting understanding. Section
\ref{Sect:Databases} presents the used databases, while implementation details and results are
presented in section \ref{Sect:Results}. The paper ends with discussions about the impact of the
results.

\section{Related Work }
\label{Sect:Related}

This work investigates the capabilities of CNNs' to recognize the subject of paintings as compared
with the performance of humans. Thus, relevant prior work refers to solutions for object and scene
recognition in paintings. As paintings are an abstraction of real images, scene recognition in
photographs is also relevant.
At last we aim to adapt information from photographs to paintings by means of style transfer.

\textbf{Object and scene recognition in paintings.} Computer based painting analysis has been in
the focus of the computer vision community for a long period. A summary of various directions
approached, algorithms and results for not-so-recent solutions can be found in the review of
Bentowska and Coddington \citeyear{Bentowska2010}. However, a plurality of works
\cite{Agarwal2015,Karayev2014,Bar2015,florea:17} addressed style (art movement) recognition as it
is the main label associated with paintings. An intermediate topic is in the work of Monroy et al.
\citeyear{Monroy2014} which detected and extracted shapes (associated with objects) but as
pre-processing for the final task which was that of restoration.

Object recognition has been in the focus of Crowley and Zisserman \citeyear{Crowley2016} while
searching the YourPaintings dataset with learning on photographic data; yet this dataset features
``older'' art, which is less abstracted than modern art.

Scene recognition in paintings is also named genre recognition following the labels from the
WikiArt collection. This topic was approached by Condorovici et al. \citeyear{Condorovici2013} and
by Agarwal et al. \citeyear{Agarwal2015};  both works, using the classical feature+classifier
approach, tested smaller databases with a few (5) classes: 500 images - \cite{Condorovici2013} and
1500 images - \cite{Agarwal2015}. More extensive evaluation, using data from WikiArt, was performed
by Saleh and Elgammal \citeyear{saleh2015}, which investigated an extensive list of visual features
and metric learning to optimize the similarity measure between paintings and respectively by Tan et
al. \citeyear{Tan:16}, which employed a fine tuned AlexNet architecture \cite{Krizhevsky:12}  to
recognize both style and genre of the paintings.

The process of transferring knowledge from natural photography to art objects has been previously
addressed beyond the recent transfer from ImageNet to WikiArt \cite{Tan:16}; yet their solution is
general as it is common to use ImageNet pre-trained CNNs on smaller databases. 3D object
reconstruction can be augmented if information from old paintings is available \cite{Aubry2013}.
Classifiers (deep CNNs) trained on real data are able to locate objects such as cars, cows and
cathedrals \cite{Crowley2016} if the artistic rendering is not very abstract. The problem of
detecting/recognizing objects in any type of data, regardless if it is real or artistic, was named
cross-depiction by Hall et al. \cite{Hall2015}; however the problem is noted as being particular
difficult and in the light of dedicated benchmarks \cite{Cai2015}, the results show plenty of space
for improvement.

A significant conclusion that arises is that all solutions that showed some degree of success
did it for \emph{older} artistic movements, where scene depiction was without particular
abstraction. To our best knowledge there is no reported significant success for (more) modern art.

\textbf{Scene recognition in photographs.} Scene recognition in natural images is an intensively
studied topic, but under the auspices of being significantly more difficult than object recognition
or image classification \cite{Zhou:14}. We will refer the reader to a recent work \cite{Heranz2016}
for the latest results on the topic. We will still note that the introduction of the SUN database
\cite{xiao2010sun} (followed by the subsequent expansions) placed a significant landmark (and
benchmark) on the issue. In this context, it was shown that using domain transfer (e.g from the
Places database), the performance may be improved \cite{Zhou:14}.

\textbf{Style transfer.} Any image claiming artistic representations adheres to some non-trivial
ideas about content, color and composition; these ideas are often grouped in \emph{styles}. It has
been an important theme in computer vision \cite{bae2006} to develop methods that are able to
transfer the style from an artistic image to a normal, common image. The methods are called
\emph{style transfer}. Currently, the methods may be classified as non-neural (such as the
Laplacian style transfer \cite{Aubry2014}) and neural artistic style transfers, initiated by the
method of Gatys et al. \citeyear{Gatys:2015}. We will detail these algorithms in section
\ref{Sect:PaintingUnderstanding}. For the moment, we will note that the efficiency of the transfer
was quantified, till now, only subjectively as that it was not rigourously defined the concept of
the style so to evaluate objectively.

\textbf{Scene recognition by humans.} While it is beyond the purpose of this paper to discuss
detailed aspects of the human neuro-mechanisms involved in scene recognition, following the
integrating work of Sewards \citeyear{Sewards2011} we stress only one important aspect: compared to
object recognition, where localized structures are used, for scene recognition the process is
significantly more tedious and complex. Object recognition "is solved in the brain via a cascade of
reflexive, largely feedforward computations that culminate in a powerful neuronal representation in
the inferior temporal cortex" \cite{DiCarlo12}. In contrast, scene recognition includes numerous
and complex areas, as the process starts with peripheral object recognition, continues with central
object recognition, activating areas such as the entorhinal cortex, hippocampus and subiculum
\cite{Sewards2011}.

Concluding, there is a consensus, from both the neuro-science and the computer vision communities
that scene recognition is a particularly difficult task. This task becomes even harder when the
subject images are heavily abstracted paintings produced within the movements of modern art.


\section{CNNs: Architectures and Training}
\label{Sect:CNN}

Following AlexNet \cite{Krizhevsky:12} performance in the ImageNet challenge, in the recent years
we have witnessed a steep increase in popularity of Convolutional Neural Networks (CNNs) when
considering the task of image classification. Especially after Donahue et al.
\citeyear{donahue2014} showed that ImageNet pre-trained CNNs provide very good descriptors, it is
very hard to find a large-enough database where state of the art performance is not related with
CNNs. As such, the bulk of our experiments revolve, in one way or another, around the CNN
algorithms, either as direct classification methods or as auxiliary tools.


The first network architecture tested in this work is AlexNet, proposed by Krizhevsky et al
\citeyear{Krizhevsky:12}. This architecture features 5 convolutional layers followed by 3 fully
connected layers. The size of the receptive fields is large ($11 \times 11$), in comparison to more
recent architectures; the design is optimized for the use of two parallel GPUs, although currently
it easily fits on a single GPU. Even though it has been visibly surpassed by other networks, it
still remains a main reference point in all architecture comparisons.

A more powerful architecture is represented by the VGG-type networks, proposed by Simonyan and
Zisserman \citeyear{Simonyan2014}. In contrast to AlexNet, this design features smaller receptive
fields ($3 \times 3$), but with a stride between regions of only 1. These decisions have led to the
possibility of increased depth, by stacking sequentially up to 5 convolutional layers, and,
subsequently, to higher performance than AlexNet. Although we do not use this network for direct
classification, the stacked convolutional layers of VGG–-19 trained on ImageNet produce very
efficient feature descriptors, and, thus, are heavily used by the style transfer algorithms
\cite{Gatys:2015,johnson2016}.

In the remainder of the paper, for the task of classification, we will use the Residual Network
(ResNet) \cite{He2016} architecture, with 34 layers. All the hyper-parameters and the training
procedure follow precisely the original ResNet \cite{He2016}. Nominally, the optimization algorithm
is 1--bit Stochastic Gradient Descent, the initialization is random (i.e. from scratch) and when
the recognition accuracy levels on the validation set, we decrease the learning rate by a factor of
10. The implementation is based on the CNTK library\footnote{Available at
\url{https://github.com/microsoft/cntk/}}.



\section{Painting Understanding and Domain Transfer}
\label{Sect:PaintingUnderstanding}

The main task of this work is to generate results about the understanding of the machine learning
systems (in our case deep CNN) grasp of art. In such a case, one needs tools to ease the
comprehension of the system internal mechanisms.

For CNNs, the most popular visualization tool has been proposed by Zeiler and Fergus
\citeyear{Zeiler:14} by introducing deconvolutional layers and visualizing activations maps onto
features. Attempts to visualize the CNNs, for scene recognition, using this technique, indicated
that activations are related to objects. Thus it lead to the conclusion that multiple object
detectors are incorporated in such a deep architecture \cite{Zhou:15}. This means that high
activations were triggered simultaneously in multiple parts of an image, being thus rather
spatially vague. In parallel, visualization of activations for genre \cite{Tan:16} have shown that,
for instance, the landscape type of scene leads to activating almost the entire image, thus being
less neat to draw any conclusion. Consequently, we tried a different approach to investigate the
intrinsic mechanisms of deep CNNs. Our approach exploits domain transfer and simply correlates the
final results with the training database.

Given the increased power of machine learning systems and the limited amount of data available to a
specific task, a plethora of transfer learning techniques appeared \cite{Lu2015}. Transfer learning
is particularly popular when associated with deep learning. First, let us recall that the lower
layers of deep nets trained on large databases are extremely powerful features when coupled with a
powerful classifier (such as SVM) and may act as feature selectors, no matter the task
\cite{donahue2014}. Secondly, the process of fine tuning deep networks assumes using a network that
has been pre-trained on another database, and, with a small learning rate, adapting it to the
current task.

In contrast, the concept of domain transfer or domain adaptation appeared as an alternative to the
increase of the amount of information over which a learner may be trained directly (without fine
tuning) in order to improve its prediction capabilities. Many previous solutions and alternatives
have been introduced. We will refer to the work of Ben-David et al. \citeyear{Ben2010} for
theoretical insights on the process.

It has been shown that domain transfer is feasible and the resulting learner has improved
performance if the two domains are adapted. In this context one needs to mention the work of Saenko
et al. \citeyear{Saenko2010} that proved that using a trained transformation, the domain transfer
is beneficial.

In our case, the target domain is the artistic painting domain, while for the source domain we
consider several alternatives: consumer photographic image domain (represented by images from the
SUN database), artistic photographic image domain (with instances from the Photo-Artist database),
and a combination of features from the painting domain (taken from the "older", less abstract
styles of paintings); these will be detailed further in section \ref{Sect:Databases} and,
respectively, in \ref{Sect:Results}.

For the domain adaptation function we investigate several alternatives. As a visual distinctive
characteristic of painting is the style, the investigated functions are in the category of style
transfer methods. As they are a key concept to our work, in the next subsection we will detail
them.


\subsection{Style transfer}
Following the seminal work of Reinhard et al. \citeyear{Reinhard:2001} the idea to transfer
elements from a reference image, with an artistic value, to a subject one increased in popularity.
The main concepts introduced then still stand: first, both images are represented into a space that
offers a meaningful description with respect to transferable elements; next, the subject image is
altered such that it fits into the description of the reference image. Quite often the second step
implies matching the estimated probability density functions (i.e histograms) of the two images. A
simple version \citeyear{Reinhard:2001,huang2017} is to assume the distribution in the feature
space to be Gaussian, independent (or at least uncorrelated) with respect to the dimensions. Such
an assumption is a sword with two edges: on one side, the feature space should be constrained to
follow the decorrelation assumption, but, on the other hand, the matching process is very simple as
only the mean and the variance on each feature dimension needs to be fitted.

While in the original work \cite{Reinhard:2001}, the envisaged transferable element is color,
later, the style of the image was approached \cite{bae2006}. Prior to popularization of CNNs, image
descriptors were based on manually--set filter weights; a prominent is the case of the Laplacian
style transfer \cite{Aubry2014}. In the last period, the spectacular results showed by Gatys et al.
\citeyear{Gatys:2015} helped the so called Neural Style Transfer methods to gain in popularity.

In the next subsections we will review the formalization of the main style transfer methods;
$\mathbf{S}$ will denote the subject image, while $\mathbf{R}$ the style reference image. In some
solutions, $\mathbf{S}$ is altered to mimic the style from $\mathbf{R}$, in others, a third image
$\mathbf{X}$, is built from scratch to match the content from $\mathbf{S}$ and the style from
$\mathbf{R}$.

\subsubsection{Laplacian style transfer}

Following the findings of Sunkavalli et al. \citeyear{sunkavalli2010}, the complex transfer process
produces better results if it is carried separately on different levels of details from the two
images. The most simple form to have various details of an image is to use a pyramidal
representation. The Laplacian style transfer \cite{Aubry2014} assumes a Laplacian pyramidal
representation with multiple levels. On each level the style transfer is implemented as gradient
transfer. Nominally, at each pixel, $\mathbf{p}$, in each level of the pyramid, given a
neighborhood $\mathbf{q}$, the set of gradients in the two images is $\nabla
S_{\mathbf{p},\mathbf{q}}$ (for the subject image) and $\nabla R_{\mathbf{p},\mathbf{q}}$ (for the
reference image). The transfer is implemented by iteratively correcting the local point
$S(\mathbf{p})$:

\begin{equation}
    \begin{array}{cl}
        \mathbf{S}_n(\mathbf{p}) &= r(\mathbf{S}_{n-1}(\mathbf{p})); \enspace
            r(i) = g + sign(i-g)t(|i - g|); \\
            t(i) &= CDF^{-1}_{[\nabla \mathbf{R}(\mathbf{p})]}\left( CDF_{[\nabla \mathbf{S}(\mathbf{p})]}(\mathbf{p}) \right)
    \end{array}
\end{equation}

where CDF is the cumulative density function computed on the image, $r(\cdot)$ is the transfer
function, while $\mathbf{S}_1(\mathbf{p})$ is the initial local pixel of the subject image and
$\mathbf{S}_n(\mathbf{p})$ is the same value at iteration $n$. The mapping needs to be done
iteratively, as in practice the CDF is approximated by the local cumulative histogram on a discrete
support of values. Typically, 10 iterations suffice.


\subsubsection{Neural style transfer algorithm} \label{Sect:NeuralStyle}

The algorithm proposed by Gatys et al. \citeyear{Gatys:2015} aims to transpose the style by
representing an image into a space where content and style are separable. This is achieved by using
the feature representation offered by the layers of a convolutional network architecture that do
not have skip connections (e.g. VGG--19). The choice for this representation originates in the
observation that higher layers (i.e. closer to the fully connected and to the classification
layers) will focus on high-level features, thus onto content (the focus is on objects, not on the
actual pixels), while the first layers can be used to reconstruct the image with accurate pixel
values.

Differently with respect to previous works which adjust the subject (content) image, the original
neural style algorithm \cite{Gatys:2015} starts from a white noise image, $\textbf{X}$, which is
modified in order to match the over–imposed statistics.

Starting from the white noise image, the method employs gradient descent to smoothly adjust values
in order to match the content and style of the desired source images. The global loss function is a
linear combination between two other loss functions used for content and, respectively, the style
matching. In the neural style transfer algorithm, the subject image will provide the content
matching, $\mathbf{S}$, while the generated image is $\mathbf{X}$. Considering the activations
$F_{ip}^l$ ($P_{ip}^l$) of the $i^{th}$ filter on the $l^{th}$ layer at position $p$ in a CNN
representation of $\mathbf{S}$ (and respectively $\mathbf{X}$), the loss function with respect to
content can be defined as:

\begin{equation}
L_{content} (\mathbf{S},\mathbf{X},l) = \frac{1}{2}\sum_{i,j} (F_{ip}^l-P_{ip}^l)^2
\end{equation}

To underline the correlations between activations, the algorithm uses the Gram matrix; $G_{ip}^l$
is the inner product between the vector versions of the $i^{th}$ and $p^{th}$ feature maps of the
$l^{th}$ layer:
\begin{equation}
G_{ip}^l = \sum_k F_{ik}^lF_{pk}^l
\end{equation}

With the Gram matrix defined, the contribution of each layer is brought into discussion. The
$l^{th}$ layer is composed of $N_l$ feature maps of size $M_l$. Thus, considering, besides
$\mathbf{X}$, the style source image $\mathbf{R}$, the Gram representations $A^l$ and $G^l$ are
computed for layer $l$. The contribution of a certain layer and the total style loss function are:

\begin{equation}
E_l=\frac{1}{4N_l^2M_l^2}\sum_{i,j}(G_{ip}^l-A_{ip}^l)^2
\end{equation}

\begin{equation}
L_{style}(\mathbf{A},\mathbf{X})=\sum_{l=0}^Lw_lE_l
\end{equation}

The total loss function, which takes into account both content and style is a linear combination of
the two aforementioned components:

\begin{equation}
L_{total}(\mathbf{X},\mathbf{S},\mathbf{R})=\alpha L_{content}(\mathbf{S},\mathbf{X})+\beta
L_{style}(\mathbf{A},\mathbf{X})
\end{equation}
where $\alpha$ and $\beta$ are manually picked weights.

The dramatic visual impact of the original visual style transfer method encouraged other works to
address the same subject. Li et al. \cite{li2017} showed that, in fact, this version of style
transfer minimizes the Maximum Mean Discrepancy (MMD) with the second order polynomial kernel thus
implementing another version of matching the feature distributions. Maximum Mean Discrepancy (MMD)
is a test statistic for the two-sample testing problem, where acceptance or rejection decisions are
made for a null hypothesis \cite{gretton2012}; thus, in other words, a minimum MMD enforces that
two distributions, in this case that of the image $\mathbf{X}$ and that associated with the
reference images ($\mathbf{S}$ for style and $\mathbf{R}$ for style) are similar enough, within a
statistical relevance limit. The last work \cite{li2017} indicates that the relation of the neural
style transfer abides to the original principle stated by Reinhard et al. \citeyear{Reinhard:2001},
namely of feature distribution matching. The neural style transfer \cite{Gatys:2015} implements an
approximation of the exact solution, which would have been given by inverting and composing the
Cumulative Density Functions (CDF).

In an attempt to accelerate the transfer, Ulyanov et al. \citeyear{ulyanov2016} and Johnson et al.
\cite{johnson2016} learned the inverse mapping by a deep CNN. Yet this approach has the
disadvantage that only a specific reference image equivalent CDF may be learned by the CNN; another
style from another reference image requires another CNN. Thus, this approximation restricts the
number of transferable styles to the number of CNNs trained. The limitation was very recently
approached by Huang and Belongie \citeyear{huang2017}, which assumes independent Gaussians and thus
match only the mean and variance (to obtain the statistical match) following by training a CNN to
approximate the multivariate Gaussian CDF and do the inverse mapping.

Although it is faster, the CNN learning of the inverse CDF as in \cite{johnson2016,ulyanov2016}
dramatically limits the number of transferable styles, making the concept ineffective as domain
adaptation method. The Gaussian matching \cite{huang2017}, while being able to transfer an infinity
of styles, still imposes two approximations that have an effect on the final image quality.
Concluding, the original neural style  transfer \cite{Gatys:2015} remains the most viable candidate
for domain adaptation between the photographic image domain and the paintings domain.

\subsection{Comparison Between Neural and Laplacian Transfers}

While the Laplacian and Neural style transfer are built on different paradigms (neural vs.
non-neural), we stress that the two methods share similar concepts. First, let us note that the
pooling paradigm from the CNN is similar with the downsampling used in the Laplacian pyramid.
Noting that the best results for the neural style transfer are achieved using a VGG-19
architecture, this implies an input image of $224\times 224$; the VGG-19 has 4 pooling layers
between convolutional layers, thus offering a 5 level pyramidal representation. For the same image
input, the Laplacian transfer uses 7 levels, thus going for coarser style too. In each such
pyramidal level, the VGG stacks 2 to 3 convolutions, while the Laplacian uses only one.

Another aspect is related to the used filters. For the Laplacian filters, the weights are preset,
always having the same values, while for the neural style transfer, the weights are learned while
trying to classify on the ImageNet task. It has been shown that the CNN lower level filters are
similar to standard gradient filters \cite{fischer2014}, yet those from upper levels have no such
correspondence. In this point, one should note that CNN architectures use multiple filters in
parallel (the VGG from 64 in the bottom layer to 512 in the upper ones), while the Laplacian only
1. This huge difference ensures that a much richer description is available for the neural
transfer.

The last aspect taken into account for the comparison is the procedure used for matching the
distributions. The original Neural Style transfer goes for minimizing the MMD, which is to ensure
that the multi-dimensional feature distributions of the image to change and of the reference image
do match from a statistical point of view. The Laplacian goes for the exact solution. Yet, the inverting of the
CDF in the Laplacian is also imperfect, as the computed ``cumulative density function'' is only an
approximation of the true CDF, aspect enhanced by the significant number of iterations, which are
required to fill the discontinuities. Another point is that the full inversion is possible because
the Laplacian uses 1D feature description, thus being less richer than the multi-dimensional one
used for the neural version.

Concluding, just by looking at the richness of the two algorithms, the neural style transfer may be
grossly perceived as a significant step ahead with respect to the Laplacian counterpart.


\section{Databases}
\label{Sect:Databases}

For the various experiments undertaken, several databases have been employed. These are either a
collection of digitized paintings, either a collection of digital photographs. In the next
paragraphs we summarize their main characteristics.

\subsection{WikiArt database}

The WikiArt is a collection of digitized paintings used to popularize art. It is in continuous
expansion, as every day new works of art are added. The version used is the same as in
\cite{Karayev2014} and it contains approximately 80,000 digitized images of fine-art paintings. We
note that this version is highly similar to ones used by prior art, to which we compare against.

The images are labelled within 27 different styles (cubism, rococo, realism, fauvism, etc.), 45
different genres (illustration, nude, abstract, portrait, landscape, marina, religious, literary,
etc.) and belong to more than 1000 artists. To our knowledge this is the largest database currently
available that contains genre annotations. For our tests we considered a set that contains 79434
images of paintings.

\begin{figure*}
 \begin{center}
  \begin{minipage}{0.8\textwidth}
    \begin{tabular}{c}
        \begin{tabular}{cccc}
            \includegraphics[width=0.228\linewidth]{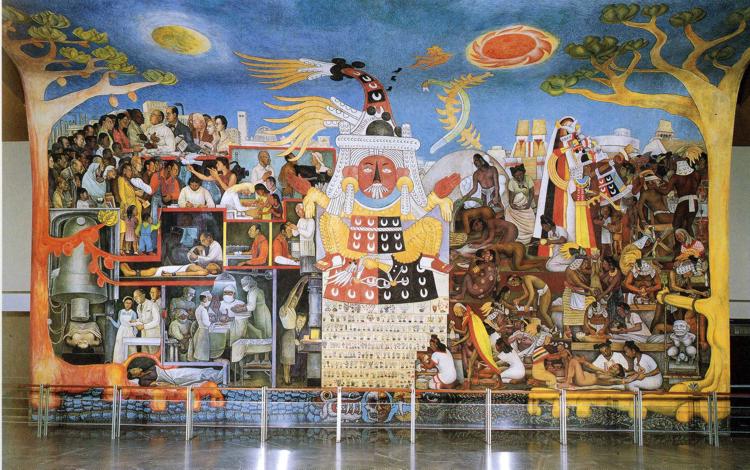}&
            \includegraphics[width=0.249\linewidth]{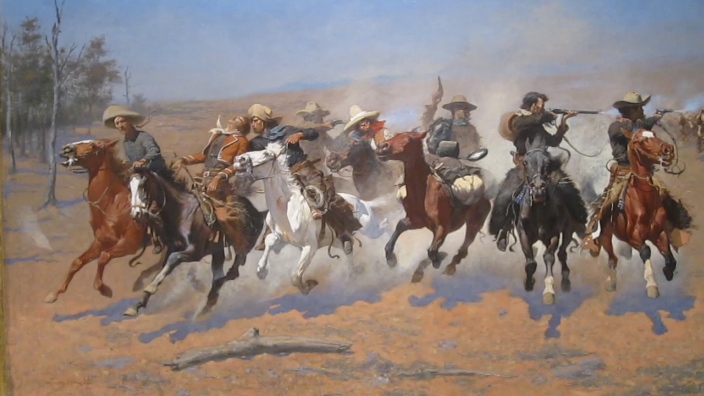}&
            \includegraphics[width=0.212\linewidth]{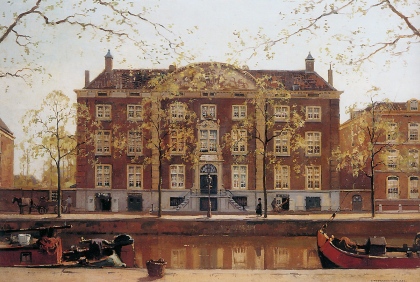}&
            \includegraphics[width=0.212\linewidth]{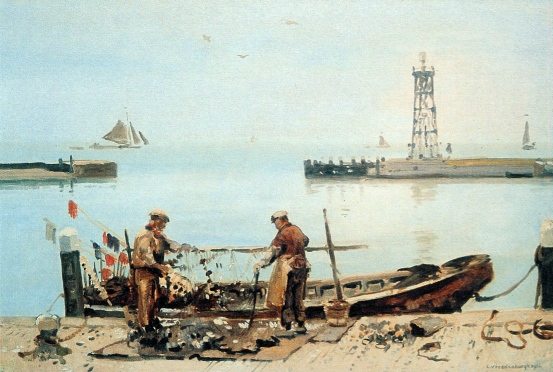}\\
            Allegorical & Battle & Cityscape  & Genre
        \end{tabular} \\

        \begin{tabular}{cccccc}
            \includegraphics[width=0.103\linewidth]{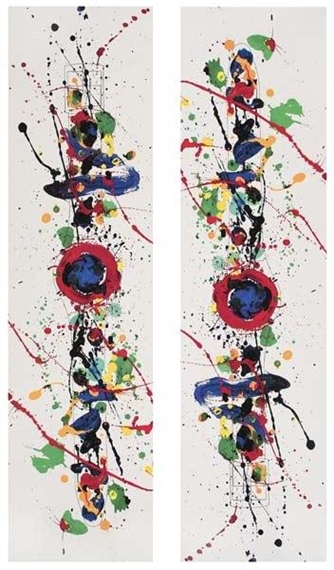}&
            \includegraphics[width=0.177\linewidth]{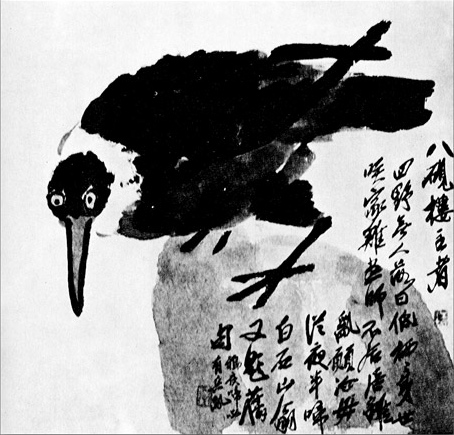}&
            \includegraphics[width=0.145\linewidth]{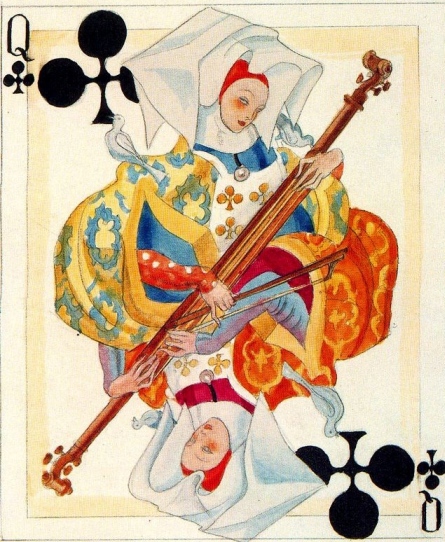}&
            \includegraphics[width=0.1398\linewidth]{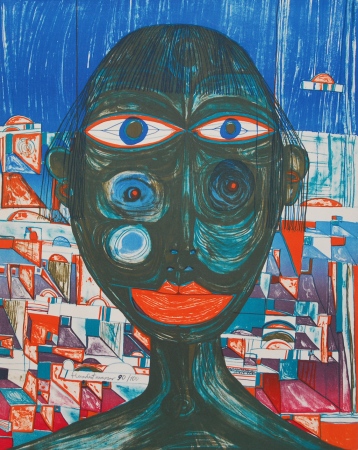}&
            \includegraphics[width=0.13\linewidth]{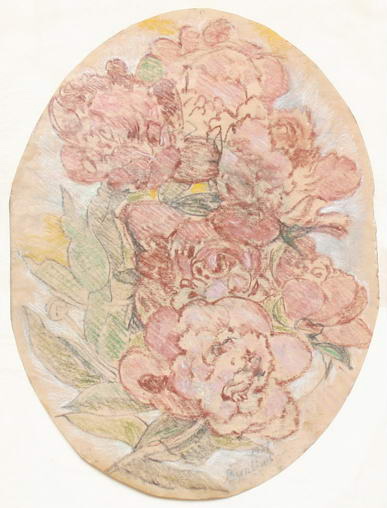}&
            \includegraphics[width=0.17\linewidth]{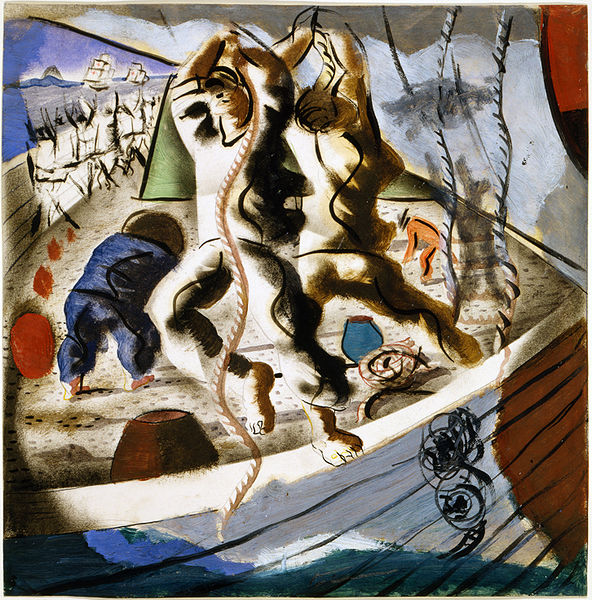}\\
            Abstract & Animal & Design & Figurative & Flower & History
        \end{tabular} \\

        \begin{tabular}{ccc ccc}
            \includegraphics[width=0.146\linewidth]{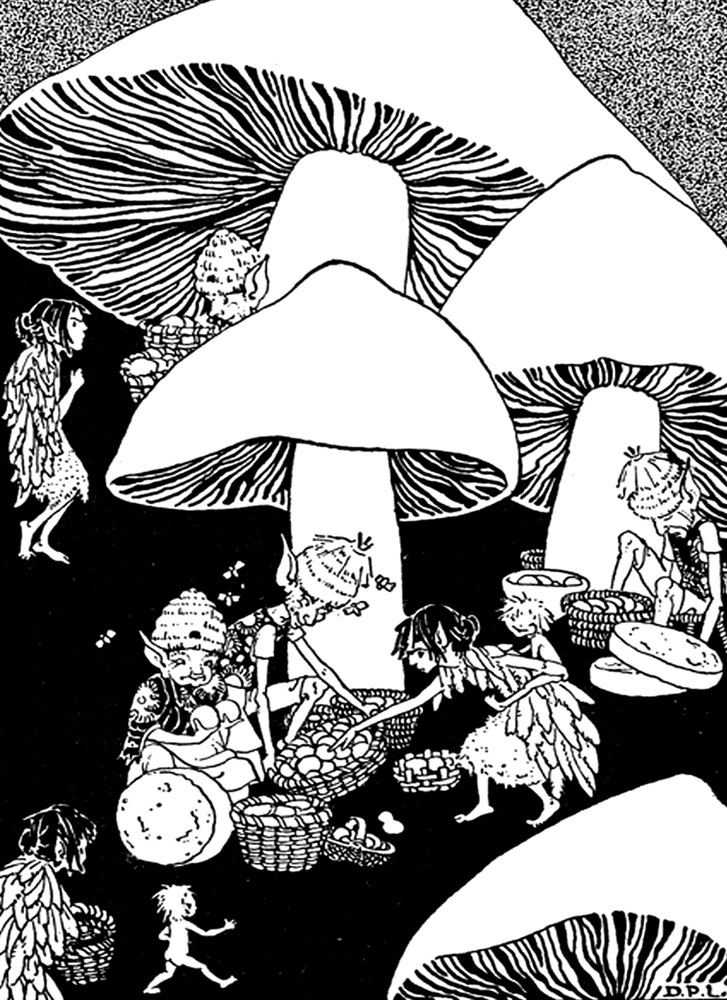}&
            \includegraphics[width=0.142\linewidth]{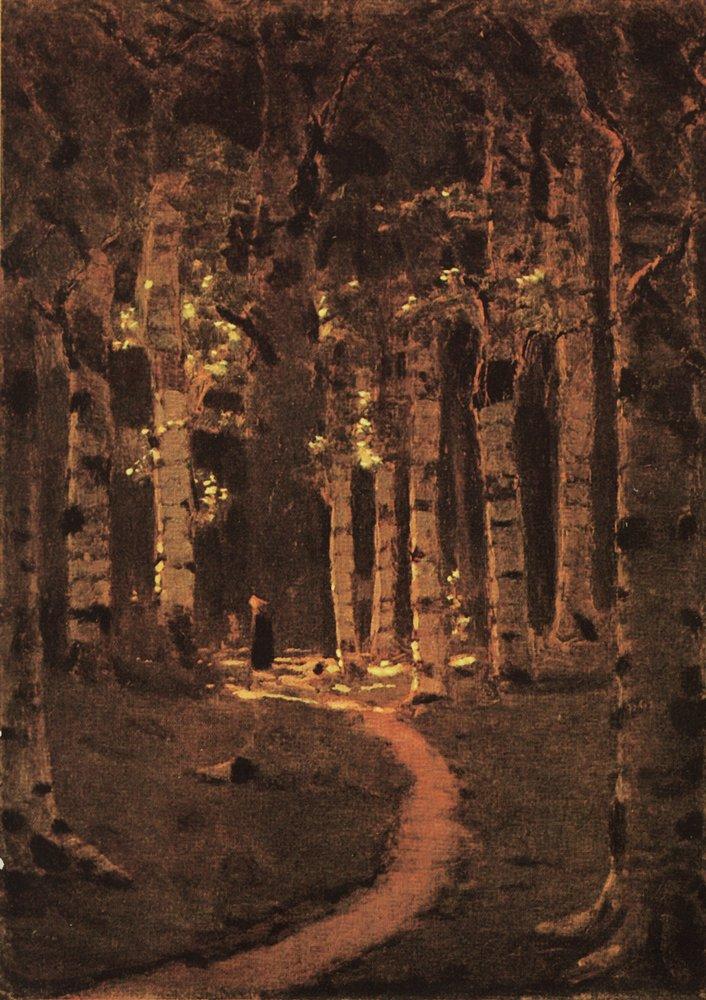}&
            \includegraphics[width=0.14\linewidth]{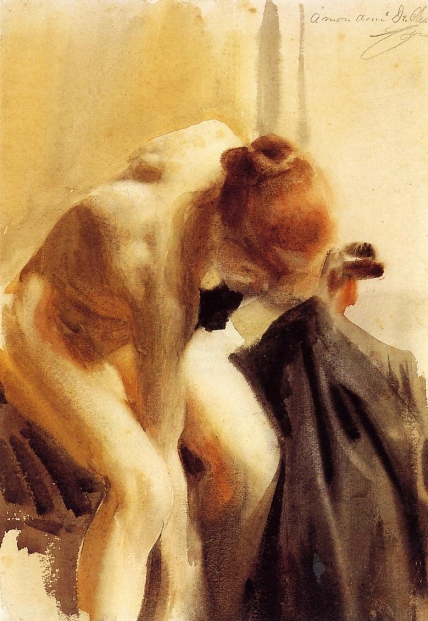}&
            \includegraphics[width=0.13\linewidth]{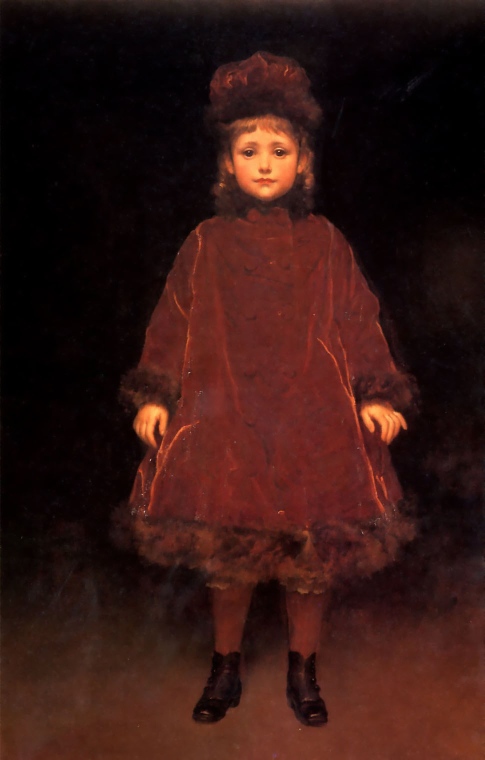}&
            \includegraphics[width=0.145\linewidth]{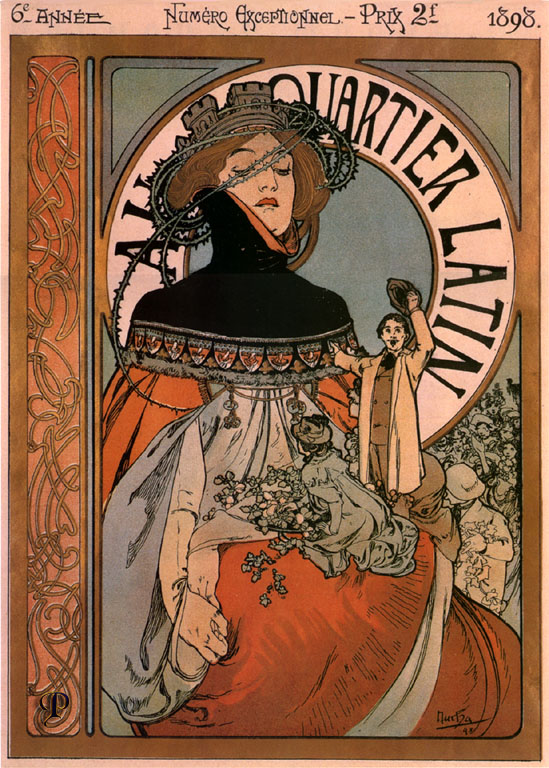}&
            \includegraphics[width=0.15\linewidth]{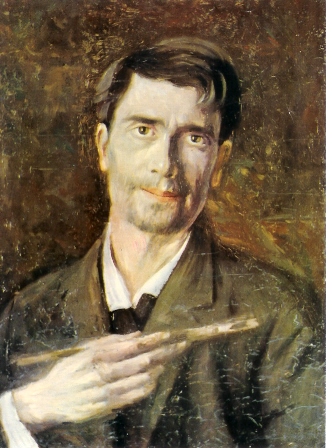}\\
            Illustration &Landscape & Nude & Portrait &Poster & Self-portrait
        \end{tabular} \\

        \begin{tabular}{cc ccc}
            \includegraphics[width=0.176\linewidth]{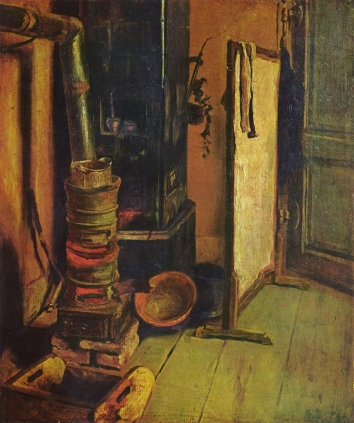}&
            \includegraphics[width=0.169\linewidth]{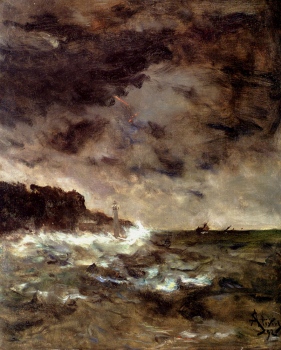}&
            \includegraphics[width=0.203\linewidth]{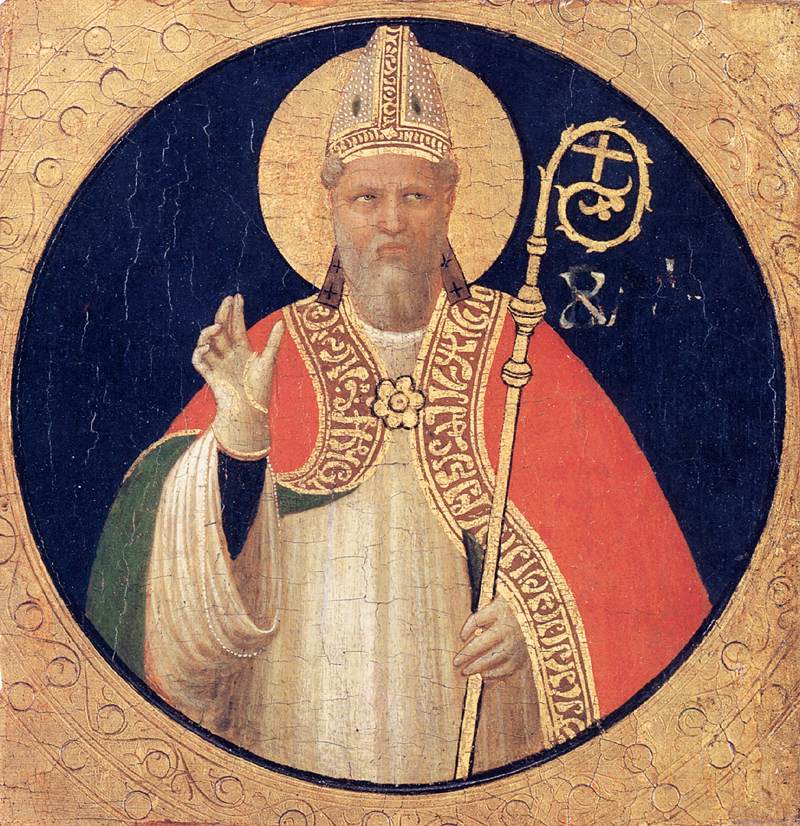}&
            \includegraphics[width=0.167\linewidth]{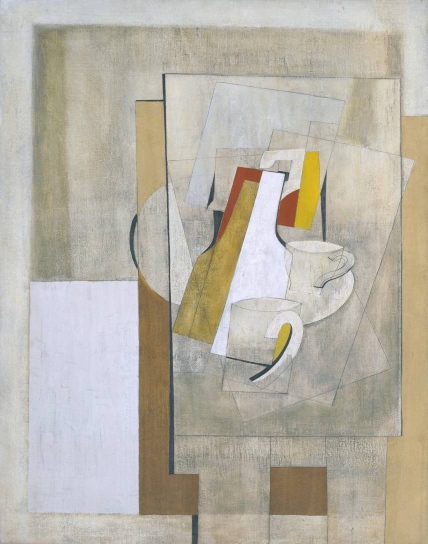}&
            \includegraphics[width=0.1545\linewidth]{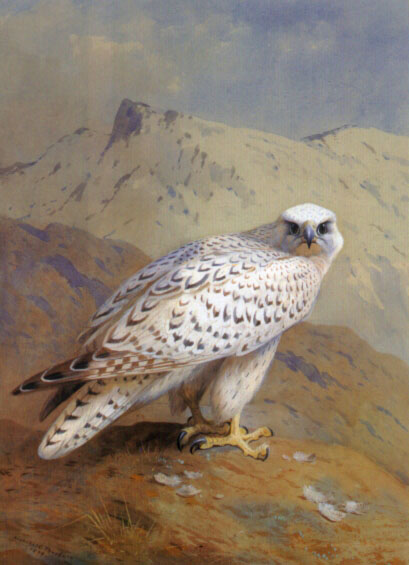}\\
            Interior & Marina  & Religious & Still & Wildlife
        \end{tabular} \\
        \begin{tabular}{cccc}
            \includegraphics[width=0.248\linewidth]{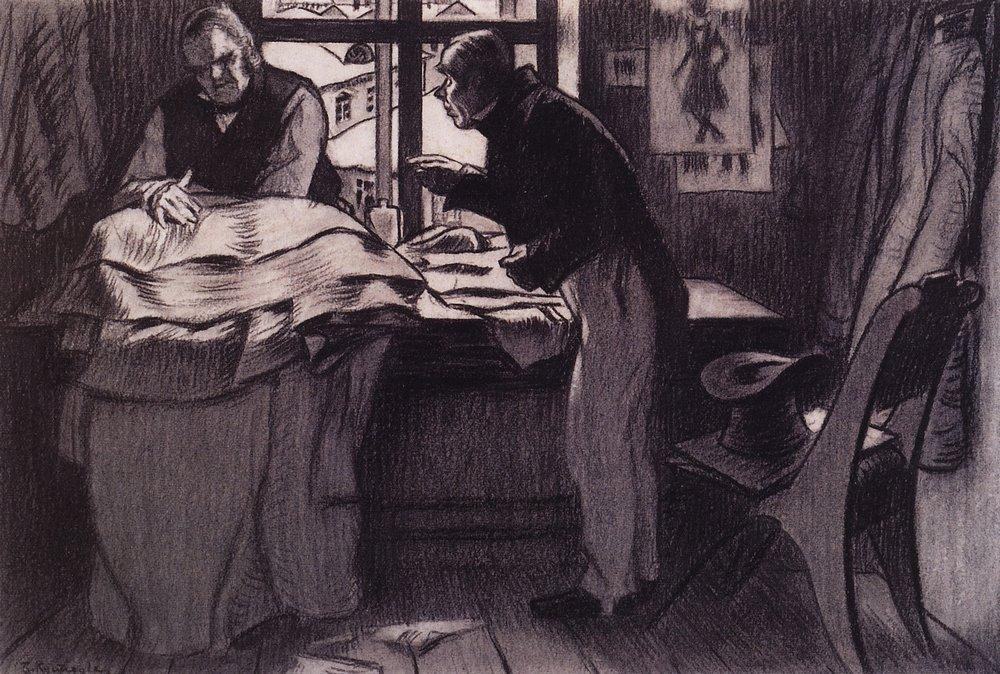}&
            \includegraphics[width=0.2025\linewidth]{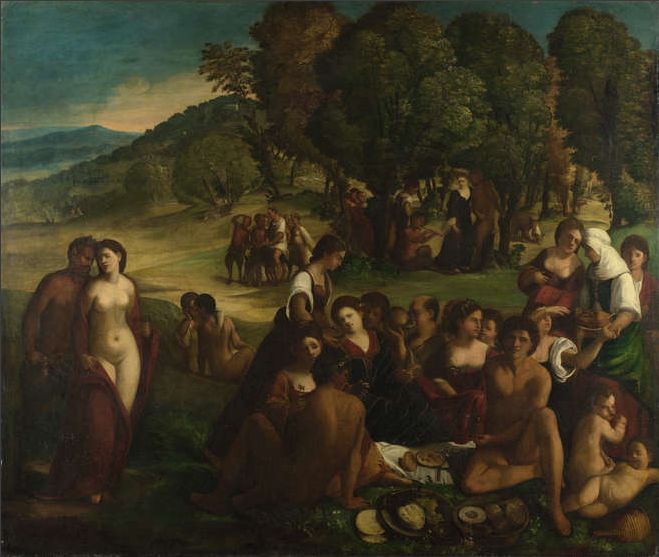}&
            \includegraphics[width=0.225\linewidth]{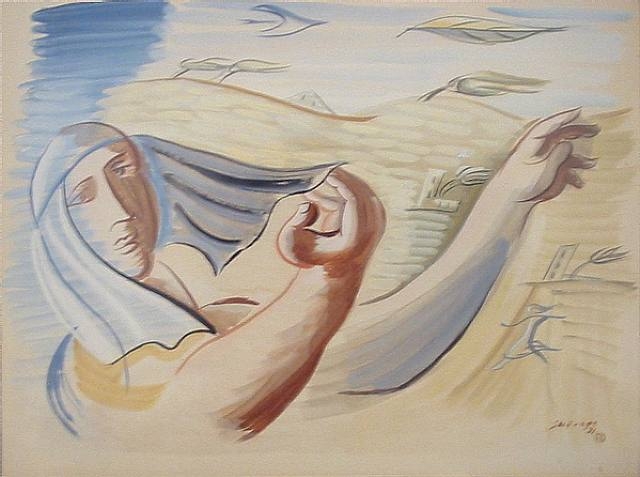}&
            \includegraphics[width=0.252\linewidth]{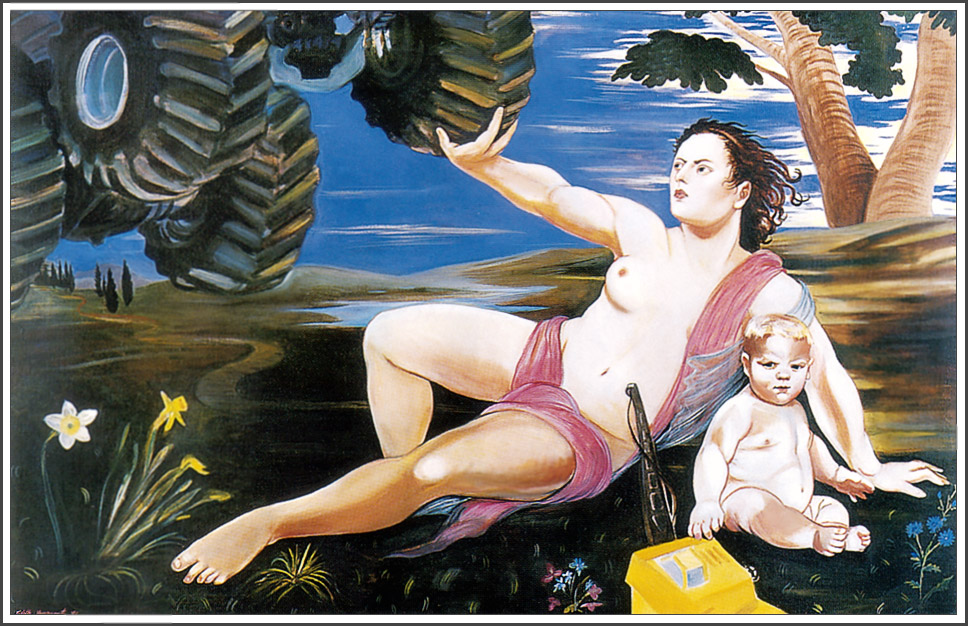}\\
            Literary & Mythological & Sketch & Symbolic
        \end{tabular} \\
    \end{tabular}
  \end{minipage}
   \caption{ Examples illustrating the 25 Genre categories  from the WikiArt collection. ``Others'' is
   not illustrated. Note that some categories can still be confused with other, such as ``Mythological'' with
   ``Symbolic'' or ``Religious''. \label{Fig:Database}}
 \end{center}
\end{figure*}

In WikiArt some of the genres were not well represented (i.e. less than 200 images) and we gathered
them into a new class called ``Others''. This led to a division of the database into 26 classes
illustrated in Figure \ref{Fig:Database}. The names of the classes and the number of training and
testing images in each class can be seen in Table \ref{Tab:Genre_Explain}. Furthermore, as the
genre label is not explained in prior works, we also contribute with an explanation of the genre
meaning; these explanations are collected from an Internet
encyclopedia\footnote{http://www.getty.edu}.

The list of genres that counted less than 200 examples and are included in the ''Others'' class
contains, among others: ''advertisement'' -- 94 examples -- illustrate a work of art designed to
market a product; ''capriccio''(194) -- depict architectural fantasies, placing together buildings
and/or elements in fantastical combinations; ``caricature''(184) - variation of sketch, but with
intended comical effect; ``cloudscape'' (162) - depicting clouds or sky; ''miniature'' (79) - small
paintings, typically originating in early middle age in the Byzantine or Central Asian empires;
''mosaic'' (17) -- images made from the assemblage of small pieces of colored materials;
''panorama''(17) --  massive artworks that reveal a wide, all-encompassing view of a particular
subject; ''pastorale'' (80) - illustrates segments of life from shepherds herding livestock around
open areas; ``quadratura'' (19) - an image based on techniques and spatial effects to create an
illusion of three-dimensional space; ``tessellation'' (122) - images of aperiodically patterns from
M. C. Escher; ``vanitas''(34) -  symbolic works of art, associated with the still life and
``veduta'' (188) -- which are a highly detailed large-scale paintings.

We note that annotation is weak, as one may find arguable labels. For instance ``literary'' and
``illustration'' categories may in fact have ``landscape'' subjects. However, as this distribution
matches practical situations, we used the database as it is, without altering the annotations.

After observing the main characteristics of paintings genres in Table \ref{Tab:Genre_Explain}, one
may conclude that the task is similar with the identification of a scene type in a photograph.
Basically, in most cases, the genre is given by the topic (subject) rendered in the artwork. Under
these circumstances, especially to specific genres, we consider that appropriate for domain
transfer, are images labelled for scene content.

\begin{table*}[tb]
\centering
 \caption{Overview of the genre organization of the WikiArt database and the explanation of
 the label meaning. We have marked only genres with more than 200 images.
 \label{Tab:Genre_Explain}}

  \begin{tabular}{|c|c|c|c|}
 \hline
    \phantom{x}  & \textbf{Genre}        & \textbf{No. imgs.} & \textbf{Description} \\ \thickhline
    1            & Abstract Art &     7201       & Uses shapes, forms etc. to replace accurate representations \\ \hline
    2            & Allegorical Painting &  809   & Expression of complex ideas using another subject \\ \hline
    3            & Animal Painting &    1233     & Paintings which depict animals \\ \hline
    4            & Battle Painting &     273     & The main subjects are battles and wars \\ \hline
    5            & Cityscape       &     4089    & Works which contain cities or other large communities \\ \hline
    6            & Design          &    1577     & Conceptual schemes of objects and structures \\ \hline
    7            & Figurative      &    1782     & Forms inspired by objective sources, but altered \\ \hline
    8            & Flower Painting &    1270     & Paintings of flowers \\ \hline
    9            & Genre Painting  &    10984     & Scenes of everyday life \\ \hline
    10           & History Painting&    656     & Depictions of historical events \\ \hline
    11           & Illustration    &    2493     & Visual representations usually meant for books, magazines, etc. \\ \hline
    12           & Interior        &    511     & Paintings depicting interiors of structures \\ \hline
    13           & Landscape       &    11548     & Contains representations of land, or other natural scenes \\ \hline
    14           & Literary Painting&   418     & Subject taken from literary work \\ \hline
    15           & Marina          &    1385    & These paintings show scenes from docks or ports \\ \hline
    16           & Mythological Painting& 1493   & Inspired by mythology \\ \hline
    17           & Nude Painting   &    1758     & Paintings which contain nudes \\ \hline
    18           & Portrait        &    12926     & Images of real individuals \\ \hline
    19           & Poster          &     229    & Works which are usually intended for advertising \\ \hline
    20           & Religious Painting &   5703   & Inspiration is drawn from religious scenes \\ \hline
    21           & Self-Portrait   &    1199     & The subject of the painting is the artist \\ \hline
    22           & Sketch and Study&    2778     & Drawings done for personal study or practice \\ \hline
    23           & Still Life      &     2464    & Images which depict inanimate objects \\ \hline
    24           & Symbolic Painting&   1959     & Content suggested by symbols in the forms, lines, shapes, and colors\\ \hline
    25           & Wildlife Painting&    259    & Paintings of natural scenes, including animals in their habitats \\ \hline
 \hline

 \end{tabular}

\end{table*}

\subsection{Natural Scenes Databases}

As a first additional source of data we have considered images from the SUN (Scene Understanding)
database \cite{xiao2010sun}. In its original form, it contains 899 classes and more than 130,000
images. Yet, only a few classes, which have a direct match with the content of the WikiArt genre
database, were selected. These classes and the number of images in each one that were added in the
training process at some step are: ``Cityscape''–-2903 images, ``Flower''–-229,
``Landscape''-–7794, ``Marina''–-1818.

Supplementary, since portraits are a very common type of paintings, we have retrieved 5993 images
from the Labelled Faces in the Wild (LFW) database \cite{Huang2007a}. Again, these images are a
mere rendering of one's face without any artistic claim. In the next section, if it is not
specified otherwise, these portraits will be associated with SUN photos. In total we have 20,075
normal photographs.

As a particular characteristic of the images from the SUN/LFW databases is that they were not
selected for their artistic value, but strictly for their adherence to one of the scene classes
tightly associated with studied genres. Artistic value, if there is any, is merely coincidental.
Also there is no abstraction in this data.


\subsection{Photo-Artist Data Set}
Thomas and Kovashka \citeyear{thomas:16} have collected a large database of artistic photographs.
Unlike the images from the SUN -- LFW database, in this case, each image has an important artistic
value, and its author is broadly accepted as an artist. The artistic value is encoded in the
composition, framing, colors, or in other elements that are harder to objectively quantify. Yet
this database was not labelled with respect to subject or scene composition.

From this large collection, we have selected and labelled a subset of 19,573 images distributed as
follows: ``Cityscape''--8068 images, ``Interior''--3036 images, ``Landscape''--4467 images,
``Portrait''--4002 images.


\subsection{Stylized Images}

The most striking visual difference between photographs and paintings is the style of representing
the subject within the paintings. In an effort to transfer knowledge between the two image types,
we have stylized some sets of images as follows:

\begin{itemize}
    \item images from the Photo-Artist database were stylized with Laplacian style transfer;
    \item images from the Photo-Artist database were stylized with neural style transfer.
    \item images of paintings with classical styles, that have a realistic depiction of reality, have been stylized
    using the neural transfer method and according to the styles of modern abstract art movements.
 \end{itemize}

\begin{figure*}[tb]
 \centering
    \begin{tabular}{c}
        \begin{tabular}{ccc}
            \includegraphics[width=0.20\textwidth]{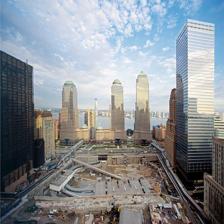} &
            \includegraphics[width=0.20\textwidth]{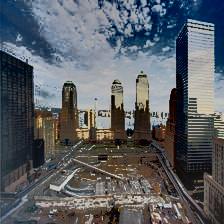} &
            \includegraphics[width=0.20\textwidth]{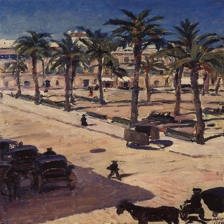}  \\
            Artistic Photo & Laplacian stylized image & Reference painting
        \end{tabular} \\
        \begin{tabular}{ccc}
            \includegraphics[width=0.20\textwidth]{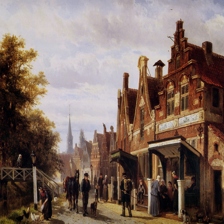} &
            \includegraphics[width=0.20\textwidth]{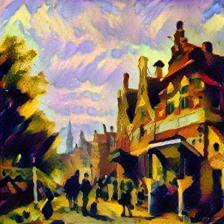} &
            \includegraphics[width=0.20\textwidth]{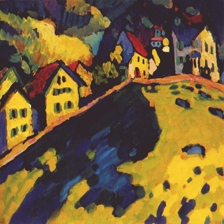}  \\
            Classic style painting & Neural stylized image & Reference painting
        \end{tabular} \\
        \begin{tabular}{ccc}
            \includegraphics[width=0.20\textwidth]{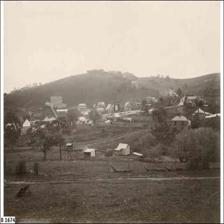} &
            \includegraphics[width=0.20\textwidth]{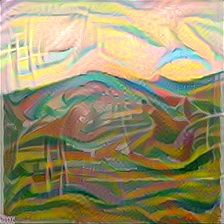} &
            \includegraphics[width=0.20\textwidth]{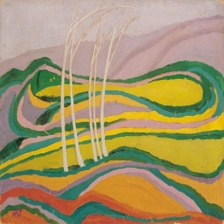}  \\
            Artistic photo & Neural stylized image & Reference painting
        \end{tabular} \\

    \end{tabular}
        \caption{Examples of the stylization. The left hand column shows the content image, the
        center column shows the stylized image, while the right hand column contains the style reference painting. In
        this case, the images are at the resolution used by the CNN ($224\times 224$). \label{Fig:Stylization}}

\end{figure*}

Example of stylized paintings may be seen in Figure \ref{Fig:Stylization}. We have not stylized
images from SUN--LFW database, as (one may see further in Table \ref{Tab:Transfer}) their
performance was not better than those of Artistic--Photo and the process is very tedious.

For the reasons previously mentioned, among neural solutions, we have used the original neural
style transfer method \cite{Gatys:2015}. The motivation for trying to transfer the abstract style
onto realistic paintings lies in the observation that, over time, the paintings adheres to the same
favorite subjects.

The process of neural stylization using the original neural solution is time consuming. For the
initial tests \cite{florea2017:Scia} we have used a NVIDIA GTX 980 Ti and a NVIDIA K40; in this
case, the stylization of an image took 2-10 minutes, which matches the more recent reports in
\cite{ChenS16f}. Further acceleration \cite{badea2017} and the use the faster NVIDIA GTX 1080 Ti
allows, currently, faster stylization with a duration scaled down to 30 sec; yet this is still slow
considering the size of the databases, so that one will not produce an extremely large quantity of
images.


\section{Implementation and Results}
\label{Sect:Results}

The main goal of this work is to study the performance of the CNNs in recognition of a paintings'
genre and, in parallel, to investigate various ways in which this performance can be increased.
This includes experiments on the classification methods themselves, in order to establish both the
state of the art performance and a baseline for further experimentation. Afterwards, we follow with
experiments on various alterations brought to the database in the context of domain transfer.

\subsection{Baseline and comparison with prior work}

Genre recognition has been previously attempted using the WikiArt collection. We show that the used
current procedure leads to slightly superior performance when compared to previously proposed
solutions. Furthermore, as the list of prior art is not too rich, we have tested some classical
solutions to establish more firmly the baseline performance.

Agarwal et al. \citeyear{Agarwal2015}, Tan et al. \citeyear{Tan:16} and Saleh and Elgammal
\citeyear{saleh2015} used the WikiArt database for training and testing in order to classify
paintings into different genres. While the first two used a very small subset, the later two
solutions focused on 10 classes from the entire database, namely: Abstract, Cityscape, Genre,
Illustration, Landscape, Nude, Portrait, Religious, Sketch and Study and Still life; in total their
subset gathered $\sim$63,000 images.

We have adopted the division (training and testing) from Karayev et al. \citeyear{Karayev2014} as
working with a complete version of WikiArt. Furthermore, we stress that in our case the images from
training and testing are completely different and are based on a random selection.

In order to compare our results to the ones reported by the mentioned previous works, we have
selected the same classes of paintings for training and testing. While in the case from
\cite{saleh2015} the test-to-train ratio is mentioned, in \cite{Tan:16}, it is not. Under these
circumstances and against our best efforts, the comparison with prior art is, maybe, less accurate.

The results from Table \ref{Table:Comp_StateOfArt} show that the proposed method gives slightly
better results than previous method \cite{Tan:16}, building upon the difference that they use a
smaller fine--tuned (i.e. initialized on ImageNet) network (AlexNet), while we have used a larger
one, based on the improved residual connections paradigm, but initialized from scratch. Also, the
previous work used database augmentation while, for this result, we have not. The difference
between here used ResNet-34 and the previously used AlexNet is above the stochastic margin, as we
will discuss later. Furthermore, we report the average performance over 5 runs. Thus we may
emphasize that our efforts produced state of the art genre recognition on the WikiArt collection.

On the 26-class problem, which will be furthered explored in the remainder of the paper, the
residual CNN is able to retrieve better performance than classical solutions. In this test, the
classical solutions coupled pyramidal Histogram of Oriented Gradients (pHog) \cite{Dalal2005} or
pyramidal Local Binary Pattern (pLBP) \cite{Ojala:2002}, as features, with a SVM for
classification. The SVM uses radial basis function and we performed the grid searched for the best
$\gamma,C$ parameters. Only features extracted from lower layers of a CNN (DeCAF) provide a
description that is competitive with CNNs. Compared to our previous work, where we have reported
59.1\% accuracy, the improved performance is due to better hyperparameter tuning. Currently, our
effort produced state of the art performance on the 26 class problem.

\begin{table*}[tb]
\centering
     \caption{Comparison with state of the art methods. The table is horizontally split  to group
     solutions that have used databases with similar size. Acronyms: BOW - Bag of Words,  ITML -
     or Iterative Metric Learning; pHoG - HoG pyramid as in \cite{Dollar14}; pLBP - LBP pyramid
     implemented in VLFeat \cite{Vedaldi:2010} DeCAF \cite{donahue2014} assumes the first 7
     levels of AlexNet trained on ImageNet. With italic we marked our proposed solution while with
     bold the top performance.
    \label{Table:Comp_StateOfArt}}
\begin{tabular}{ |c|c|c| c|c|c | c|} \thickhline

\textbf{Method} & \textbf{No. classes} & \textbf{No. images} & \textbf{Test ratio} &
\textbf{Accuracy} (\%)
\\ \thickhline
    Agarwal et al. \citeyear{Agarwal2015} - SIFT+BOW & \multirow{2}{*}{5} & \multirow{2}{*}{1500} & \multirow{2}{*}{10\%} & 82.53 \\ \cline{1-1} \cline{5-5}
    Agarwal et al. \citeyear{Agarwal2015} - ensemble &  &  &  & 84.46 \\ \hline \hline

    Saleh and Elgammal \citeyear{saleh2015} - Classemes+Boost& \multirow{6}{*}{10} & \multirow{6}{*}{63.691} & \multirow{3}{*}{33\%} & 57.87 \\ \cline{1-1} \cline{5-5} 
    Saleh and Elgammal \citeyear{saleh2015} - Classemes+ITML&   &   &  & 60.28 \\  \cline{1-1} \cline{5-5}
    Saleh and Elgammal \citeyear{saleh2015} - Classemes+Fusion& &   &  & 60.28 \\   \cline{1-1} \cline{4-5}

    Tan et al. \citeyear{Tan:16} AlexNet - scratch &   &   & \multirow{2}{*}{n/a}& 69.29 \\ \cline{1-1} \cline{5-5}

    Tan et al. \citeyear{Tan:16} CNN- finetune &   &   &   & \emph{74.14} \\ \cline{1-1} \cline{4-5}

     \emph{Proposed - ResNet 34 - scratch} &   &  & 20\% & \textbf{75.58}  \\ \hline \hline

         pHoG + SVM          &  \multirow{6}{*}{26}  & \multirow{6}{*}{79,434} & \multirow{6}{*}{20\%} & 44.37 \\ \cline{1-1} \cline{5-5}
         pLBP + SVM          &                       &  & & 39.58 \\ \cline{1-1} \cline{5-5}
         DeCAF + SVM         &                       &  & & 59.05 \\ \cline{1-1} \cline{5-5}
         AlexNet - scratch        &                    &  & & 53.02 \\ \cline{1-1} \cline{5-5}
         ResNet 34 - scratch \cite{florea2017:Scia} &     &  &  & 59.1  \\ \cline{1-1} \cline{5-5}
    \emph{Proposed - ResNet 34 - scratch} &     &  &  & \textbf{61.64}  \\ \cline{1-1} \cline{5-5}
    \emph{Proposed - ResNet 34 - scratch + augmentation} &     &  &  & \textbf{63.58}  \\ \thickhline

\end{tabular}
\end{table*}

\textbf{Database Basic Augmentation.} Besides using domain transfer and domain adaptation, one may
improve the recognition performance by adding the original images modified by basic image
processing techniques. The processing methods that have produced positive effects are flipping and
slight rotation. The flipped versions are all horizontal flips of the original images. Regarding
the rotations, all images have been rotated either clockwise or counterclockwise with $3^0$, $6^0$,
$9^0$ or $12^0$. Adding noise, slight tone adjustment or small translations did not help.

When the size of the training database doubled, the improvement in recognition performance was of
2\%, reaching $63.58\%$  while doubling the training time too. We consider the increase too small,
so in the remainder of the experiments we have used only the original paintings without basic
augmentation.

\begin{table}[t]
\centering
 \caption{Top--$K$ accuracy for genre recognition rate in the 26-case scenario. Top--$K$ means that a positive recognition is
   achieved for an example, when any of the any the CNN model $K$ highest probability answers
   match the expected answer.\label{Tab:TopN}}
    \begin{tabular}{|c|c|c |c|c|c|}
    \hline
     \textbf{K}  & \textbf{1} & \textbf{2} & \textbf{3} & \textbf{4} & \textbf{5} \\ \hline
     Recogn.[\%] & 61.64 & 76.24 & 82.14 & 88.46 & 90.09 \\ \hline
    \end{tabular}
\end{table}

\begin{figure}[t]
 \centering
    \includegraphics[width=0.85\linewidth]{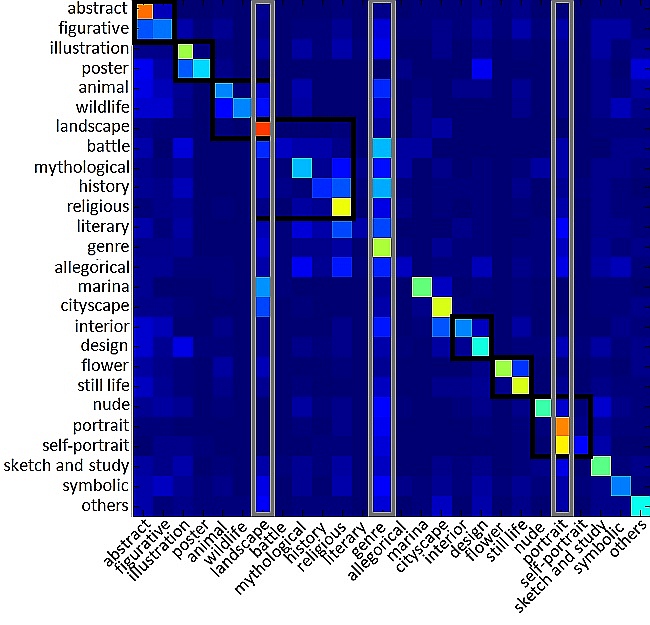}
        \caption{The confusion matrix for the 26 class. We have grouped and marked confusion classes.}
\label{Fig:Confusion}
\end{figure}

\textbf{Confusion Matrix}. The confusion matrix for the best performer on the 26-class experiment
is presented in Figure \ref{Fig:Confusion}. We have marked classes that are particularly confused.
It should be noted that from a human point of view, there is certain confusion between similar
genres such as historical$\leftrightarrow$battle$\leftrightarrow$religious,
portrait$\leftrightarrow$self portrait, poster$\leftrightarrow$illustration,
animal$\leftrightarrow$wildlife etc. Some of these confusable images are, in fact, shown in Figure
\ref{Fig:Database}.

Consequently, the Top--$K$ error may also show relevance, as in many cases there are multiple genre
labels that can be truthfully associated with one image. The performance with respect to $K$ is
presented in table \ref{Tab:TopN}. One may note that greater improvement is from Top--1 to Top--2.
For the best proposed alternative, ResNet with 34 layers, the Top-5 error is 9.91\% - corresponding
to \textbf{90.09}\% accuracy. For the 10-class experiment the Top--5 accuracy is 96.75\%.

\textbf{Stochastic Effect}. This  test studied the effect of the stochastic nature of  the CNN.
Factors such as the random initialization of all the parameters can influence the results of any
considered network. We ran the ResNet--34 several (5) times on the 26 classes and the accuracy
results had a mean of 61.64\% (Top-–1 accuracy) and a standard deviation of 0.33\%. The results
underline the fact that even though there is some variation caused by randomness, it does not
influence the system significantly.

On the other hand, if alternative solutions show variation smaller than \textbf{0.33}\%
(``stochastic margin'' as we named it), one may argue that these are not relevant to draw
conclusions.

For the following experiments we will refer solely to the 26 classes scenario, as it is the most
complete. We recall that the baseline performance (and also the best) is \textbf{61.64\%}.


\subsection{Influence of the artistic style }

Prior art \cite{Zhou:15} suggests that even in the case of the scene rendered in photographs, in
fact, a deep network builds object detectors and can recognize objects presented in \emph{a way
seen before}. These conclusion are based on the viewing method of the CNN filters based on
deconvolution \cite{Zeiler:14}. In the case of realistic scenes, Zhou et al. \citeyear{Zhou:15}
showed that filters activation is grouped on certain objects in association with a certain class.
In the case of paintings' genre, Tan et al. \citeyear{Tan:16}, using the same visualization
technique, showed much more sparsity.

In the same, the deconvolution method has no procedure to identify the failure; in other words
there is no way to prove that nothing in particular influences the decision on one class, but
rather the small, easy--to--neglect weights, exploited in Distillation -- Dark Knowledge
\cite{hinton2015}. To avoid this potential uncertainty, our experiments focus directly on the
network output, when the training set was adjusted into a specific direction.

With regard to the artistic style, we have performed two experiments. In the first one, we separate
the training and testing based on style, thus asking the CNN to generalize across style, while in
the second we have reviewed the baseline performance with respect to style.

For the first experiment, considering the full 79,434 images genre database, we selected all the
images that are associated with \emph{Cubist} and \emph{Naive Art} styles and placed them in
testing, resulting in 4,132 images for evaluation and 75,302 for training. Although numerically
this is a weaker test than the baseline, as the training set is larger, the results are
considerably worse: \textbf{50.82}\% Top--1 accuracy and \textbf{82.10}\% Top--5. We consider that
this drop (from 61.64\% - Top--1 and 90.99\% Top--5) is due to the fact that these particular
styles are rather different from the rest and the learner had no similar examples in the training
database. Also these results argue for \emph{a style oriented domain adaptation}.

\begin{figure*}
\centering
    \includegraphics[width=0.65\linewidth]{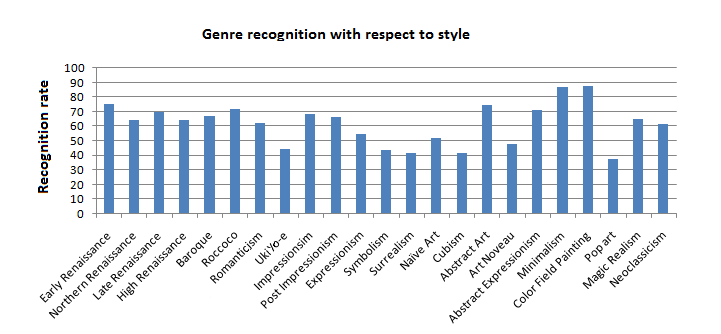}
    \caption{The genre recognition rate with respect to style. We represented the style in an
    approximate chronological order of their appearance. \label{Fig:Style_recognition}}
\end{figure*}

The second experiment benefits from the fact that WikiArt images are annotated with multiple labels
categories, and one of these category is the style. Thus given the baseline framework (i.e. keeping
the same training and testing sets), we retrieve the genre recognition rate with respect to style
(artistic movement). The results are showed in Figure \ref{Fig:Style_recognition}.

One may note that better performance is achieved for older styles, from the early beginning of
artistic painting (as we understand it today) to Post-Impressionism; these styles have a clear
representation without too much abstraction. These results are in concordance with previous object
detection in paintings \cite{Crowley2016}, where good performance is achieved for paintings
depicting scenes in a rather classical and realistic manner.  Also, we achieve good performance for
modern styles where the subject is rather unique; in cases such as Minimalism, Abstract Art or
Color Field Painting the genre is in fact the style.

In contrast, for styles such as Surrealism, Naive Art, Cubism, Pop Art, the degree of abstraction
is high and the CNN has difficulties in interpreting the scene; the scene subject is also variable.
Based on this observation, coupled with the observation that in many cases modern paintings
reinterpret classical compositions  under the rules of new styles, we attempted to create new
pseudo-paintings by using a realistic depiction of an old content and a modern style; this is the
reason while we tried to neurally transfer abstract styles on older paintings content. Furthermore,
this idea is in the same line with neural transfer methods
\cite{Gatys:2015,johnson2016,ulyanov2016,huang2017}, where almost every time, the reference image
had an abstract style, while content that easily comprehensible.


\subsection{Domain transfer}

The domain transfer experiments were performed iteratively, starting from small sets to larger
ones. The motivation lies in the fact that producing relevant images for transfer requires
non-negligible effort. We recall that we annotated $\sim 20,000$ artistic photos and we produced
more that $30,000$ neurally adapted images.

Each iteration assumes $N$ ($N=250, 500 \dots$) maximum images in each of the 26 classes; on some
classes, since they have too few examples, even in the first iteration this maximum is too generous
from the begining. For a detailed number of images existing in each class,  we kindly ask the
reader to look into Table \ref{Tab:Genre_Explain} and to keep in mind that 80\% is used for
training while 20\% is for testing. For iterations corresponding to given $N=250,\dots, 10000$, we
have added, in all domain transfer scenarios, images from three classes, as follows: cityscape --
2903, landscape -- 4467, portrait -- 4002. At the end, we have added all images from all classes
and all the images available in alternative, transferable, domains.

\begin{figure}
\centering
    \includegraphics[width=0.88\linewidth]{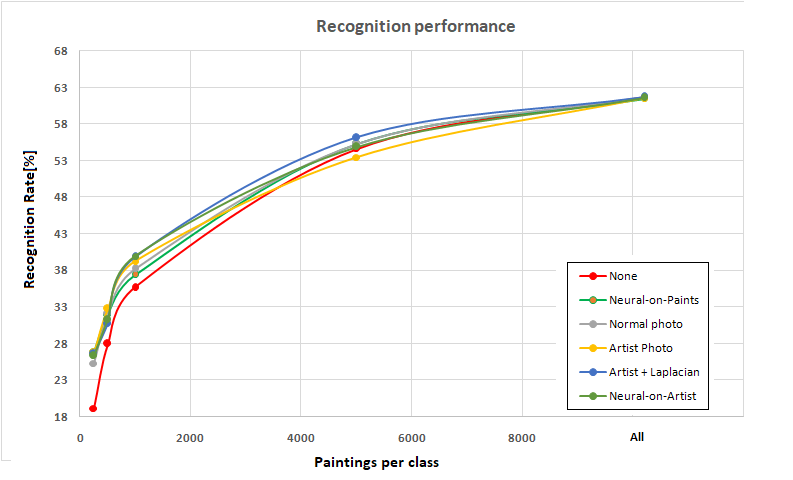}
    \caption{The genre recognition rate with respect to transferred images per class, as a function
    of the augmentation method. When the number of transferred images
     dominate over the number of painting in the database, the performance improves  \label{Fig:Domain_Transfer}}
\end{figure}

The obtained results may be numerically followed in Table \ref{Tab:Transfer}, while visually in
Figure \ref{Fig:Domain_Transfer}. The baseline performance, where no domain transfer is involved,
is on the row marked with ``None''. The ``Best-improvement'' row marks the difference between the
best achieving solution and the baseline. ``Added image ratio'' row represents the ratio between
the number of added images and the number of paintings used in this training. Each cell from the
column marked with ``Avg. improv.'' (average improvement) indicates the mean of the improvement for
the transfer method.

\begin{table*}[tb]
\begin{center}
    \caption{Recognition rates[\%] when adding images from ``Cityscape'', ``Portrait'', ``Landscape'' to the paintings.
    At each iteration we keep only the nominated value (250, 500, etc.)
    from each of the 26 classes. We add the non-paintings from the three classes originating in some domain
    or adapted with some function. On each column (testing scenario) we marked with bold letters the
    best performance. Row ``None`` indicates the performance when no transfer was performed and thus gives
    the baseline. ``Best-improvement'' shows the difference, per column, between the best performance
    and baseline. ``Added image ratio'' indicates the percentage of images added to the baseline.
    Column ``Avg. improv.'' is the improvement of a transfer scenario with respect to baseline.
    \label{Tab:Transfer}}

\begin{tabular}{|c| c|c|c|  c|c||c|}
\thickhline
    \multirow{2}{*}{Method}  & \multicolumn{5}{|c|}{\textbf{Recognition rates [\%] when paintings per class}} & \textbf{Avg.}\\ \cline{2-6}
    \phantom{x} & \textbf{250} & \textbf{500} & \textbf{1000} & \textbf{5000} &  \textbf{All}& \textbf{Improv.}\\ \thickhline
    None                    & 19.12 & 27.95 & 35.66 & 54.61  & 61.64 & 0 \\ \hline \hline
    Normal Photos (SUN/LFW) & 25.28 & 32.04 & 38.21 & 55.31  & 61.67 & 2.71 \\ \hline
    Artist Photos           & 26.72 & \textbf{32.75} & 39.27 & 53.49 & 61.55 & 2.96\\ \hline
    Laplacian-on-Artist     & 26.56 & 30.73 & {39.84} & \textbf{56.17} & \textbf{61.73} & 3.21 \\ \hline
    Neural-on-Paints        & \textbf{26.76} & 31.98 & 37.4  & 55.33 & 61.47 & 2.79 \\ \hline
    Neural-on-Artist        & 26.33 & 31.27  & \textbf{39.94} & 54.88 & 61.62 & 3.01 \\ \hline \hline
    Best-improvement        & 7.64  & 4.8    & 4.18  &  1.56  & 0.09 & -- \\ \hline \hline
    Added image ratio[\%]   & 190.04& 103.38 & 58.98 & 27.33  & $\sim$26& -- \\ \hline
\end{tabular}
    \vspace{0.1cm}
\end{center}
\end{table*}

One should note that improvement (larger than stochastic margin) is achieved only when the number
of added images is with at least 30\% larger than the number of paintings in the training database
and the number of paintings is not too large. Looking at the trends from the last two rows, one may
estimate that in order to achieve some $5-10\%$ increase for the entire database, one will need
about 3 times the number of paintings to be obtained from some other source, while the same
procedure is assumed.

Another visible trend is that original paintings are better that any transferred and adapted image
from other domain. We explain this behavior by the \emph{content originality} of the paintings:
artistic works should be considerably different in at least one aspect (e.g. content, local style,
composition, subject, etc.) from its predecessors, as to be acknowledged as art. This produces a
naturally \emph{sparse} \emph{domain} representation for paintings. The sparsity is not easily
filled up by knowledge transfer from a similar representation.

At last, the performance of any domain transfer method is not too different from any other.
Although by looking at overall improvement, the Laplacian style transfer on artistic photos, is
marginally better, the difference is too small (especially given the stochastic margin) that one
cannot conclude about a clear winner. This results is somehow surprising, as the highly praised
neural style transfer (which produces visually pleasant images) does not show better performance
than the Laplacian transfer or than the images themselves.

When comparing the normal photos with artistic photos, the artistic content and the way this
content is represented, gives slightly better results. Yet a CNN learns too little from older
artistic content to deal with new one.

When comparing the Laplacian transfer with the Neural style transfer, as introduced by Gatys et al.
\citeyear{Gatys:2015}, the significant better representation (larger depth of the feature map) and
the better fitting of the latter achieves nothing. At the end of this analysis, we must note that,
although claimed, none of these methods implements a ``style transfer'', where ``style'' has the
meaning under which is used in art. Although this result has a negative connotation in the sense
that expected hypothesis (``style transfer methods does not transfer styles``) does not hold, we
feel that we should make it public, especially given the recent argumentation by Borji
\citeyear{Borji2018}. Furthermore, we see that has a strong positive connotation too: it shows that
it is much easier to improve the performance of the CNN when dealing with understanding art by
simply showing it very large quantities of relevant photographs.

\section{Discussion and Conclusions}

In this paper we discussed the CNN capabilities to recognize the scene (genre) in a painting. The
first contribution is that we clearly showed that machine learning systems (deep CNNs) are confused
by the abstraction level from art. The experiment with abstract art showed that they cannot easily
generalize with respect to style and that, the more abstract a style is, the lower is the
recognition rate of the genre (painting subject). In this sense, the CNN is similar with humans,
who also find the abstract representations of scenes to be more puzzling.

The secondary set of contributions results from the experimentation with domain transfer as an
alternative to increase the overall performance and we have found that: (1) limited improvement is
easily doable when the training set of paintings is small; (2) methods claiming to transfer style,
either Laplacian based, either neural based, although produce visually very pleasant and intriguing
images, are ineffective as domain adaptation methods with respect to style. One possible
explanation is that ``style'' as understood by the transfer methods is not the same with ``artistic
style'' in the sense of art movement.

Lastly, the third contribution is related to understanding the structure of the paintings domain.
Due to the necessary criteria for some work to be accepted as a work of art, which implies
significant artistic novelty with respect to its predecessors, the paintings domain is more sparse
than that of the normal images. Given the problem to improve the performance, a CNN learns to
better deal with some new work of art, when more works of art are presented to it in the training
set. The CNNs are similar to humans in this behavior as well, as art expert do not learn their job
looking at normal images.

\section{Acknowledgment}

The work was supported by grants of the Romanian National Authority for Scientific Research and
Innovation, CNCS UEFISCDI, number PN-II-RU-TE-2014-4-0733 and respectively, CCCDI-UEFISCDI, project
number 96BM. The authors would like to thank NVIDIA Corporation for donating the Tesla K40c GPU
that helped run the experimental setup for this research.

\bibliographystyle{icml2014}
\bibliography{SceneRecPaint}

\end{document}